\documentclass[10pt,twocolumn,letterpaper]{article}

\usepackage[pagenumbers]{iccv} 
\usepackage{multirow}
\usepackage{amssymb}
\usepackage[normalem]{ulem} 
\usepackage{float}

\definecolor{iccvblue}{rgb}{0.21,0.49,0.74}
\usepackage[pagebackref,breaklinks,colorlinks,allcolors=iccvblue]{hyperref}

\title{3R-GS: Best Practice in Optimizing Camera Poses Along with 3DGS} 

\author{
Zhisheng Huang$^{1}$ 
\and 
Peng Wang$^{2}$ 
\and 
Jingdong Zhang$^{1}$ 
\and 
Yuan Liu$^{3}$
\and
Xin Li$^{1}$ 
\vspace{0.01cm}
\and 
Wenping Wang$^{1}$
\and
\\
$^1$Texas A\&M University, $^2$Hong Kong University, $^3$Hong Kong University of Science and Technology
}

\let\oldtwocolumn\twocolumn
\renewcommand\twocolumn[1][]{%
   \oldtwocolumn[{#1}{
    \setlength{\abovecaptionskip}{0.cm}
    \begin{center}
    \centering
    \includegraphics[width=\textwidth]{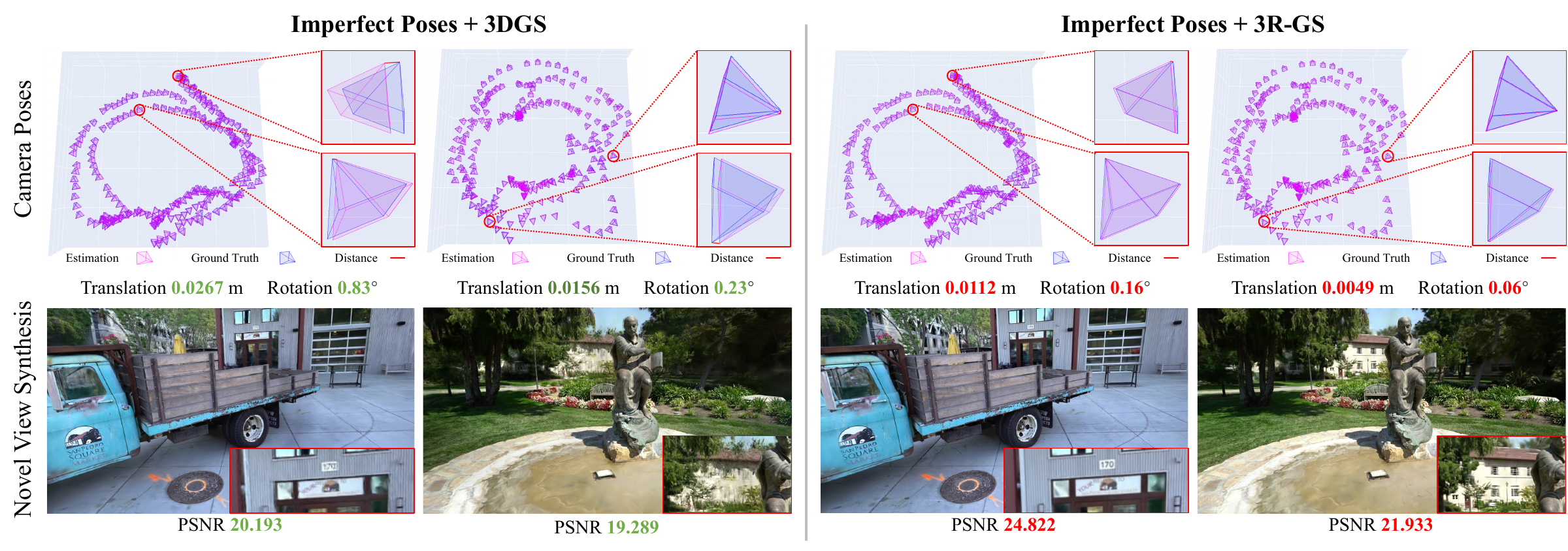}
    \vspace{-0.2em}
    \vspace{-0.4cm}
    \captionof{figure}{We propose 3R-GS, a robust method for reconstructing high-quality 3D Gaussians and poses from the MASt3R's imperfect output cameras. Our method outperforms simply joint camera pose optimization along with 3DGS in a large margin.}
    \label{fig:teaser}
\end{center}
}]
}

\begin{document}
\maketitle
\vspace{-6em}
\begin{abstract}
3D Gaussian Splatting (3DGS) has revolutionized neural rendering with its efficiency and quality, but like many novel view synthesis methods, it heavily depends on accurate camera poses from Structure-from-Motion (SfM) systems.
Although recent SfM pipelines have made impressive progress, questions remain about how to further improve both their robust performance in challenging conditions (e.g., textureless scenes) and the precision of camera parameter estimation simultaneously.
We present 3R-GS, a 3D Gaussian Splatting framework that bridges this gap by jointly optimizing 3D Gaussians and camera parameters from large reconstruction priors MASt3R-SfM.
We note that naively performing joint 3D Gaussian and camera optimization faces two challenges: the sensitivity to the quality of SfM initialization, and its limited capacity for global optimization, leading to suboptimal reconstruction results.
Our 3R-GS, overcomes these issues by incorporating optimized practices, enabling robust scene reconstruction even with imperfect camera registration.
Extensive experiments demonstrate that 3R-GS delivers high-quality novel view synthesis and precise camera pose estimation while remaining computationally efficient.
Project page: \url{https://zsh523.github.io/3R-GS/}
\end{abstract}
\vspace{-2em}    
\section{Introduction}
\label{sec:intro}

Building a 3D representation from 2D images has been a long-standing research challenge over recent decades. Recently, methods such as Neural Radiance Fields (NeRF)~\cite{mildenhall2021nerf} and 3D Gaussian Splatting (3DGS)~\cite{kerbl3Dgaussians} have emerged as powerful approaches for 3D scene representation, particularly for novel view synthesis.
These NeRF- and 3DGS-based methods require accurate camera parameters to correctly establish the 3D-2D projection relationship, a task that typically relies on structure-from-motion (SfM) techniques (e.g., COLMAP)~\cite{schonberger2016structure}.
However, SfM processes are often time-consuming and sometimes are not robust for scenes or objects with featureless regions, such as indoor environments — making them less reliable.

Recent advances in learning-based, feed-forward dense reconstruction methods (e.g., DUSt3R~\cite{wang2024dust3r} and MAST3R~\cite{duisterhof2024mast3r}) have demonstrated the significant potential of large models for inferring 3D structures from uncalibrated images.~\cite{yang2025fast3r, duisterhof2024mast3r, elflein2025light3r}.
They are more robust than traditional SfM pipelines, especially under challenging conditions such as pure rotational motions, textureless regions, and sparse view scenarios.
Despite these improvements, the estimated camera poses of these 3R-based methods still lack perfect accuracy due either to limitations in the feed-forward paradigm or to oversimplified global optimization strategies. Consequently, these imperfections can degrade the performance of subsequent 3DGS training, which requires camera poses with pixel-level accuracy.

In this paper, we introduce 3R-GS, a robust method for reconstructing high-quality 3D Gaussian representations from imperfect outputs of the MASt3R camera.
Our method builds on the idea of simultaneously learning camera poses and 3D Gaussian representations.
However, directly applying a joint learning approach for 3DGS on imperfectly registered camera frames poses several challenges.

The first challenge is \textbf{sensitivity to initialization}.
Training 3DGS requires advanced engineering heuristics, such as split/clone and opacity resetting, which require extensive hyperparameter tuning. When camera poses are imperfect, joint optimization can easily get trapped in local minima. 
This issue becomes more severe with recent feed-forward dense reconstruction methods, such as DUSt3R~\cite{wang2024dust3r}. These methods often generate point clouds with low accuracy in background regions due to high depth ambiguity present in the training data. 

The second challenge is \textbf{inefficient pose optimization}.
Unlike NeRF-like methods, which render using ray marching with pixel-level precision, 3DGS lacks built-in mechanisms for efficiently optimizing multiple cameras in a single training step. 
Instead, 3DGS uses a differentiable rasterizer that, in each training step, transforms all points into the same Normalized Device Coordinates (NDC) space, then projects, sorts, and renders them onto the image space, producing full-image level rendering.
While this mechanism renders a large batch of pixels at once, all pixels originate from a single camera, meaning that the optimization affects only one camera per training step.
In contrast, NeRF-like methods allow more efficient camera optimization since each training step can incorporate pixels rendered from multiple cameras. 

To address the first issue,
we propose adopting 3DGS-MCMC~\cite{kheradmand20253d} to enhance robustness against imperfect initialization. 
We view 3D Gaussians as MCMC samples drawn from a distribution that accurately represents the scene.
Through state transitions, the Gaussian primitives are relocated, helping them escape local minima and improving convergence. This reduces the method's  reliance on high-quality point cloud and camera pose initialization. 
In addition, with 3DGS-MCMC, we eliminate the need for heuristic densification and pruning strategies in 3DGS, removing the burden of hyperparameter fine-tuning. 

To address the second issue, drawing inspiration from PoRF~\cite{bian2023porf} and ACE0~\cite{brachmann2024scene}, we improve camera pose optimization by modeling correlations between camera poses using a multilayer perceptron (MLP). Specifically, we jointly train a globally shared MLP alongside per-camera embeddings to refine camera poses.
Additionally, to further enhance camera pose optimization, we incorporate an epipolar distance loss as a geometric constraint for refining camera poses.
This approach directly leverages pairwise correspondences image matching to optimize camera poses.
By leveraging all available pairwise correspondences, we can better optimize camera poses using more direct geometric supervising signals.
To the best of our knowledge, our approach is the first to apply MLP pose modeling and epipolar loss to tackle the unique challenges of joint 3DGS and camera pose optimization.

In summary, our contributions are:
\begin{enumerate}
    \item We propose 3R-GS, a robust method for reconstructing high-quality 3D Gaussians and poses from the MASt3R's imperfect output cameras. 
    \item Identifying two main challenges in bundle-adjusting 3DGS, we propose an effective solution that combines 3DGS-MCMC, an MLP-based pose refiner, and an epipolar distance loss to address these issues.
    \item Our experiments demonstrate the superior performance of 3R-GS in both novel view synthesis and camera pose estimation.
\end{enumerate}

\section{Related Work}

\subsection{Camera Pose Estimation from Images}

Estimating camera poses robustly and accurately using only RGB images is a long-standing and fundamental challenge.
Depending on whether the images are captured in an ordered or unordered sequence, methods such as Structure-from-Motion (SfM) or Simultaneous Localization and Mapping (SLAM) can be employed to achieve precise pose estimation.

In this paper, we primarily focus on unordered settings, as exemplified by structure-from-motion methods. Traditional SfM method, such as COLMAP~\cite{schonberger2016structure}, is a long pipeline compromising several stages like feature matching, camera registration, and bundle adjustment, is complex and not robust to some challenging cases like texture-less regions and pure rotations.
Recent learning-based sparse feature extraction and matching methods~\cite{sarlin2020superglue, lindenberger2023lightglue, tyszkiewicz2020disk, potje2024xfeat} have sought to improve traditional feature matching, yet their robustness still leaves room for improvement.
Unlike complex pipelines, recent methods—such as DUSt3R~\cite{wang2024dust3r} and its variants~\cite{duisterhof2024mast3r, wang20243d, wang2025continuous} have demonstrated the power of directly predicting 3D structures using large models.
However, despite their robustness, these approaches lack pixel-level accuracy, which leads to suboptimal downstream reconstruction results.

\subsection{Novel View Synthesis (NVS)}

Novel view synthesis, as its name suggests, aims to generate images from unseen viewpoints, leveraging input images.
This capability is pivotal in applications like virtual reality, telepresence, etc.
In recent years, Neural Radiance Fields (NeRF)~\cite{mildenhall2021nerf} and 3D Gaussian Splatting (3DGS)~\cite{kerbl20233d} have greatly improve the quality of novel view synthesis, achieving photo-realistic results.
In particular, 3DGS-like methods have been at the forefront of NVS research recently due to their clear, explicit representation and real-time rendering capabilities.
Several variants of 3DGS have been proposed, each targeting a specific aspect of the problem—for example, approaches for large-scale reconstruction~\cite{kerbl2024hierarchical,ren2024octree,zhou2024drivinggaussian,liu2024citygaussian,liu2024citygaussian,yan2024street}, feed-forward models~\cite{zhang2024gs, xu2024grm, tang2024lgm, lu2025matrix3d}, surface reconstruction~\cite{huang20242d, zhang2024rade, yu2024gaussian, yu2025gsdf}, and methods for handling reflective objects~\cite{jiang2024gaussianshader, zhang2024ref, ye20243d}.
However, these NVS methods often depend on dense, accurate camera poses obtained through SfM pipelines.
When the input camera poses are inaccurate, misalignments and artifacts can occur in the synthesized views.

\subsection{Joint NVS and Pose Estimation}

To address the challenge of reliance on the known camera poses as described above, recent methods have integrated pose optimization, designing end-to-end frameworks for joint camera pose estimation and NVS. For example, Guo et al. \cite{guo2021novel} proposed a two-stage network that synthesizes novel views directly from a 6-DoF camera pose, decoupling geometric mapping and texture rendering to enhance robustness against variable operating conditions.

Early works \cite{wang2021NeRF, lin2021barf, yen2021iNeRF, chng2022gaussian, chen2023local, jeong2021self, bian2023nope, truong2023sparf} on NeRF try to eliminate such requirement. 
Among them, NeRFmm \cite{wang2021NeRF} demonstrates the joint optimization of camera parameters and NeRF parameters through an empirical, two-stage pipeline.
BARF \cite{lin2021barf}, in contrast, introduces a single course of coarse-to fine registration on coordinate-based scene representation. 
GARF \cite{chng2022gaussian} and \cite{xia2022siNeRF} employ special activation functions, alleviating issues with high-frequency positional encoding and systematic sub-optimality in NeRFmm respectively.
NoPe-NeRF \cite{bian2023nope} adopts additional single view depth estimation to provide strong geometry cues. 
SPARF \cite{truong2023sparf}, SC-NeRF \cite{jeong2021self} and PoRF \cite{bian2023porf} incorporate image correspondence in the joint optimization.
Though impressive results have been achieved, these methods are limited to either forward-facing scenes or short video clips with simple trajectories.
Moreover, NeRF representation makes these methods slow to converge.

Recent studies has shifted focus from NeRF to 3DGS, as it enables real-time rendering, improved rendering quality and faster training speed.
In the scope of joint NVS and pose estimation, CF-3DGS \cite{fu2024colmap} and \cite{sun2024correspondence} assume a sequential video frame inputs and processes frames in a sequential manner, progressively training the 3DGS.
InstantSplat \cite{fan2024instantsplat} leverages DUSt3R \cite{wang2024dust3r} for camera pose initialization, but limited to very few images.
ZeroGS \cite{chen2024zerogs} utilizes a pretrained model as nueral scene representation, enabling training 3DGS from hundreds of unposed and unordered images. However, they also features progressive training and need a two-stage strategy for convergence.
BAD-Gaussian \cite{zhao2024bad} and \cite{darmon2024robust} also consider camera optimization in training 3DGS, but focus on addressing motion blur.

Our work is largely inspired by pioneering works on NeRF. 
For instance, the MLP pose refiner has been introduced by PoRF and ACE0 \cite{brachmann2024scene}, and epipolar distance loss function has also been utilized by SC-NeRF and PoRF.
However, different from them, we use these approaches to effectively handle unique problems in bundle adjusting 3DGS.
And compared progressive methods in 3DGS, our methods only need to modify the standard 3DGS training pipeline slightly with negligible overhead and can be applied to both short video clips and full video sequences.

\section{Method}
\begin{figure*}[t]
    \centering
    \vspace{ -0.6cm}
    \includegraphics[width=\textwidth]{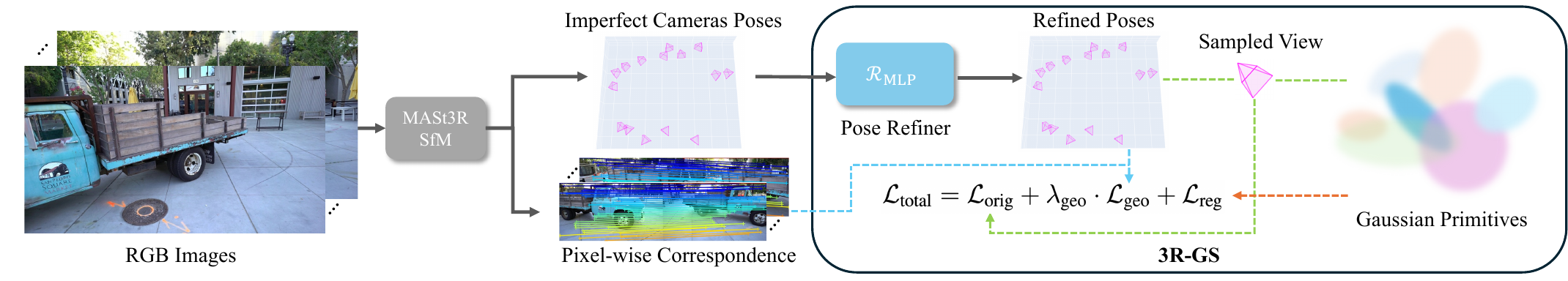}
    \vspace{-0.7cm}
    \caption{Overview of the 3R-GS pipeline. The pipeline jointly refines camera poses and 3D Gaussian parameters.}
    \vspace{-0.6cm}
    \label{fig:pipe}
\end{figure*}

\subsection{Overview}
Given a set of images captured without known poses in challenging scenes, our goal is to reconstruct both high-quality 3D Gaussian representations and accurate camera poses.
Existing 3D Gaussian reconstruction methods rely heavily on accurate camera poses—typically obtained from traditional structure-from-motion techniques (e.g., COLMAP~\cite{chen2022structure}) as input, and often struggle in scenarios such as textureless indoor environments.

To this end, we build on recent techniques that incorporate large reconstruction priors~\cite{wang2024dust3r, duisterhof2024mast3r}, and specifically, we employ MASt3R-SfM~\cite{duisterhof2024mast3r} to robustly estimate camera poses.

While MASt3R-SfM outperforms traditional SfM methods such as COLMAP~\cite{schonberger2016structure} in terms of robustness under various conditions, its estimated camera poses remain imperfect due to a lack of pixel-level accuracy, posing challenges for the downstream 3DGS reconstruction.
Our 3R-GS, a joint 3DGS and camera poses learning framework from MASt3R-SfM, aims to address the above issue.
However, naively optimizing the imperfect camera poses during 3DGS training leads to only limited improvements, introducing two challenges - sensitivity to initialization and inefficient pose optimization, as described in the introduction.

\begin{figure}[t]
\includegraphics[width=\linewidth]{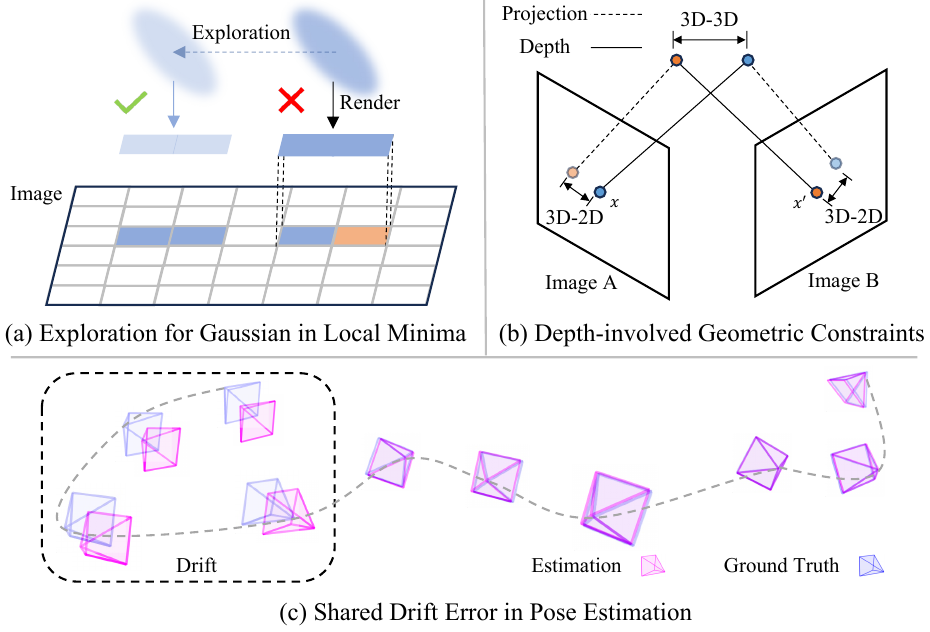}
\vspace{-0.6cm}
\caption{Motivations for 3R-GS; see Sec.~\ref{sec:mcmc}, \ref{sec:epipolar}, and \ref{sec:mlp-refiner}.}
\label{fig:bags}
\vspace{-0.6cm}
\end{figure}

To address these challenges, we introduce:
(1) a robust pose refinement strategy leveraging Markov Chain Monte Carlo (Section.~\ref{sec:mcmc}),
(2) a global camera correlation model using an MLP-based refiner (Section.~\ref{sec:mlp-refiner}), and
(3) a rendering-free geometric constraint based on epipolar loss (Section.~\ref{sec:epipolar}).

\subsection{MCMC-based Pose Optimization}
\label{sec:mcmc}
\noindent\textbf{Motivation:}
Vanilla 3DGS optimization is highly sensitive to initialization because Gaussian primitives have limited adaptability and rely on accurate initial point clouds \cite{rota2024revising, kheradmand20253d}.
For example, as shown in Fig.~\ref{fig:bags}(a), if a Gaussian primitive is initially placed slightly away from its ideal position due to imperfect initialization, it may struggle to correct itself. This happens because the photometric rendering loss only provides gradients within a small local region, making it hard for the primitive to escape local optima and reach the correct position. As a result, poor initialization can lead to suboptimal convergence and degraded scene reconstruction.

Furthermore, adaptive density control in 3DGS depends on gradient magnitude-based thresholds, which require manual tuning or adjustments to the densification strategy \cite{du2024mvgs} when introducing new training objectives. This reliance not only complicates optimization but also poses additional challenges when jointly optimizing camera poses with 3DGS.

\noindent\textbf{Solution:} 
We adopt 3DGS-MCMC \cite{kheradmand20253d}, which improves robustness to initialization by reformulating 3D Gaussian Splatting as Markov Chain Monte Carlo (MCMC) sampling. This approach interprets training as sampling from a distribution \( p(\mathcal{G}) \) that assigns high probability to collections of Gaussians faithfully reconstructing training images. It reveals that standard 3DGS optimization resembles Stochastic Gradient Langevin Dynamics (SGLD) updates:  
\[
\mathcal{G} \leftarrow \mathcal{G} + a \cdot \nabla_{\mathcal{G}} \log p(\mathcal{G}) + b \cdot \eta
\]
where \( \eta \) is exploration noise, and parameters \( a \) and \( b \) balance convergence and exploration.
With this noise, exploration in Fig.~\ref{fig:bags}(a) can be achieved.
Moreover, 3DGS-MCMC removes the need for heuristic-based densification and pruning by replacing them with principled state transitions. We also incorporate their regularizer to promote parsimonious use of Gaussians.

With 3DGS-MCMC, we achieve robust joint optimization of camera poses and 3DGS, which addresses the ``sensitivity to initialization'' issue. In the following sections, we introduce two techniques to tackle the ``inefficient pose optimization'' challenge.

\subsection{MLP-Based Global Pose Refinement}
\label{sec:mlp-refiner}
\noindent\textbf{Motivation:}
In practice, multiple cameras often share common drift errors — while their relative poses may be correct, they collectively deviate from the ground truth with a shared rotation and translation error as shown in Fig.~\ref{fig:bags}(c). 
However, directly optimizing individual camera poses treats them as independent, which can distort originally correct local relative poses and make the optimization more prone to local minima due to the inherent non-convexity of the problem~\cite{pan2024global}.

\noindent\textbf{Solution:} We introduce an MLP-based global pose refiner, which learns to predict pose corrections $\Delta \mathbf{T}_i$ from a latent camera representation:
\begin{equation}
    \Delta \mathbf{T}_i = \mathcal{R}_{\text{MLP}}(\mathbf{z}_i),
\end{equation}
where $\mathbf{z}_i$ is a learnable camera embedding jointly optimized with the MLP refiner. The corrections consist of translation ($\Delta \mathbf{t}_i \in \mathbb{R}^3$) and rotation ($\Delta \mathbf{r}_i \in \mathbb{R}^6$) components. The MLP is initialized with a zero-mean prior to ensure stable refinement. This formulation captures global pose relationships by using a shared MLP across all views, enabling more accurate camera adjustment. In practice, it achieves significantly better results than directly optimizing individual camera poses.

\subsection{Rendering-Free Geometric Constraint}
\label{sec:epipolar}
\noindent\textbf{Motivation:} 
In addition to the factor previously mentioned that contributes to inefficient camera optimization, another issue is that relying solely on rendering loss lacks direct geometric supervision for camera poses.
A straightforward approach to imposing direct geometric supervision is to use correspondence-based geometric losses.
We note that MASt3R-SfM provides matching correspondences, which can be potentially used for our geometric optimization.
Specifically, MASt3R-SfM constructs a a sparse scene graph $\mathcal{G} = (\mathcal{V}, \mathcal{E})$, where each vertex $ I \in \mathcal{V} $ represents an image, and each edge $ e = (n, m) \in \mathcal{E} $ denotes an undirected connection between two likely overlapping images $ I^n $ and $ I^m $.
Based on the graph, MASt3R-SfM computes correspondence matches $ \mathcal{M}^{n, m}$.

To utilize the correspondences, common choices include the 3D-2D projection loss \cite{truong2023sparf, duisterhof2024mast3r} and 3D-3D loss \cite{chen2022structure, duisterhof2024mast3r}, both relying on depth as shown in Fig.~\ref{fig:bags}(b).
The 3D-3D loss computes distances between corresponding points in an image pair by back-projecting them to 3D space using depth and camera parameters; and 3D-2D loss re-projects these 3D points onto the image plane to compute 2D distances to their correspondences. These methods usually require multiple image pairs to simulate global bundle adjustment, ensuring more consistent gradients.
However, integrating these optimization targets into 3DGS training presents significant challenges, primarily because 3DGS employs per-view depth sorting for rendering (both RGB and depth) and requires tens of thousands of iterations for training. This computational constraint severely limits the number of views that can be processed in each step. Incorporating additional views substantially increases training time and memory consumption to prohibitive levels. 
Consequently, only a subset of image pairs can be considered in each step when applying the aforementioned geometric constraints, preventing the enforcement of a truly global objective and ultimately leading to suboptimal results.

\noindent\textbf{Solution:} 
We propose a rendering-free global geometric constraint based on epipolar distances. Given image correspondences $\mathcal{M}^{n,m}$ from MASt3R-SfM, we define the loss as:
\begin{equation}
\mathcal{L}_{\text{geo}}=\frac{1}{|\mathcal{E}|} \sum_{(n, m) \in \mathcal{E}} \frac{1}{|\mathcal{M}^{n, m}|} \sum_{(x_i, x_i^{\prime}) \in \mathcal{M}^{n, m}}
 \operatorname{conf}_i \cdot d\left(x_i, x_i^{\prime}\right)
 \label{eq:geo}
\end{equation}
where $conf_i$ is confidence provided by MASt3R for correspondence $(x,x')$, and $d(x, x')$ is the symmetric epipolar distance computed from the fundamental matrix $F$, which is derived from the camera poses and intrinsics.
Unlike PoRF \cite{bian2023porf}, we consider the correspondences from all image pairs \((n, m) \in \mathcal{E}\) in MASt3R-SfM during each training iteration. This enables a more globally informed joint optimization of camera pose. While MASt3R-SfM can provide thousands of correspondences for each image pair, we empirically find that only a few hundred are necessary, so we subsample the correspondences uniformly.

\subsection{Final Training Objective}
\label{sec:to}
Our complete training objective integrates the original 3D Gaussian Splatting rendering losses with additional regularization terms from 3DGS-MCMC \cite{kheradmand20253d}, along with our geometric constraints $\mathcal{L}_{\text{geo}}$ in Eq.~\ref{eq:geo}.

\begin{equation}
    \mathcal{L}_{\text{total}} = \mathcal{L}_{\text{orig}} + \lambda_{\text{geo}} \cdot \mathcal{L}_{\text{geo}} + \mathcal{L}_{\text{reg}}
\label{eq:total}
\end{equation}

The original 3DGS training loss $\mathcal{L}_{\text{orig}}$ follows \cite{kerbl3Dgaussians}:
\begin{equation}
    \mathcal{L}_{\text{orig}} = (1 - \lambda_{\text{D-SSIM}}) \cdot \mathcal{L}_{1} + \lambda_{\text{D-SSIM}} \cdot \mathcal{L}_{\text{D-SSIM}}
\end{equation}
where $\mathcal{L}_{1}$ measures L1 color error and $\mathcal{L}_{\text{D-SSIM}}$ measures structural similarity, with $\lambda_{\text{D-SSIM}}=0.2$.
The regularization term $\mathcal{L}_{\text{reg}}$ in 3DGS-MCMC  promotes efficient use of Gaussians by encouraging fewer Gaussians:
\begin{equation}
    \mathcal{L}_{\text{reg}} = \lambda_{o} \cdot \sum_{i}|o_i|_1 + \lambda_{\Sigma} \cdot \sum_{ij}|\sqrt{\text{eig}_j(\Sigma_i)}|_1
\end{equation}
where $\text{eig}_j(.)$ denotes the $j$-th eigenvalue of the covariance matrix.
\section{Experiments}

\begin{table*}
    \centering
    \vspace{-0.3cm}
    \resizebox{\textwidth}{!}{ 
\begin{tabular}{cccclccclccclccclccc}\hline
                         & \multicolumn{3}{c}{3DGS}                                                                      &  & \multicolumn{3}{c}{Spann3R}                                                                    &  & \multicolumn{3}{c}{ZeroGS}                                                                    &  & \multicolumn{3}{c}{CF-3DGS}                                                                   &  & \multicolumn{3}{c}{Ours}                                                                      \\ \cline{2-4} \cline{6-8} \cline{10-12} \cline{14-16} \cline{18-20}\multirow{-2}{*}{Scenes} & PSNR$\uparrow$                          & SSIM$\uparrow$                          & LPIPS$\downarrow$                         &  & PSNR$\uparrow$                           & SSIM$\uparrow$                          & LPIPS$\downarrow$                         &  & PSNR$\uparrow$                          & SSIM$\uparrow$                          & LPIPS$\downarrow$                         &  & PSNR$\uparrow$                          & SSIM$\uparrow$                          & LPIPS$\downarrow$                         &  & PSNR$\uparrow$                          & SSIM$\uparrow$                          & LPIPS$\downarrow$                         \\ \hline
Truck                    & \cellcolor[HTML]{FFCE93}20.91 & \cellcolor[HTML]{FFCE93}0.723 & \cellcolor[HTML]{FFCE93}0.181 &  & \cellcolor[HTML]{FFFC9E}10.67  & \cellcolor[HTML]{FFFC9E}0.398 & \cellcolor[HTML]{FFFC9E}0.863 &  & -                             & -                             & -                             &  & -                             & -                             & -                             &  & \cellcolor[HTML]{FFCCC9}24.82 & \cellcolor[HTML]{FFCCC9}0.860 & \cellcolor[HTML]{FFCCC9}0.121 \\
Ignatius                 & \cellcolor[HTML]{FFFC9E}18.96 & \cellcolor[HTML]{FFCE93}0.665 & \cellcolor[HTML]{FFFC9E}0.249 &  & 13.32                          & \cellcolor[HTML]{FFFC9E}0.298 & 0.589                         &  & \cellcolor[HTML]{FFCCC9}21.95 & \cellcolor[HTML]{FFCE93}0.665 & \cellcolor[HTML]{FFCE93}0.234 &  & -                             & -                             & -                             &  & \cellcolor[HTML]{FFCE93}21.93 & \cellcolor[HTML]{FFCCC9}0.778 & \cellcolor[HTML]{FFCCC9}0.198 \\
Caterpillar               & \cellcolor[HTML]{FFCE93}19.29 & \cellcolor[HTML]{FFCE93}0.539 & \cellcolor[HTML]{FFCE93}0.349 &  & \cellcolor[HTML]{FFFC9E}12.57  & \cellcolor[HTML]{FFFC9E}0.348 & \cellcolor[HTML]{FFFC9E}0.720  &  & -                             & -                             & -                             &  & 12.96                         & 0.340                         & 0.616                         &  & \cellcolor[HTML]{FFCCC9}23.37 & \cellcolor[HTML]{FFCCC9}0.773 & \cellcolor[HTML]{FFCCC9}0.235 \\
Meetingroom              & \cellcolor[HTML]{FFCE93}22.78 & \cellcolor[HTML]{FFCE93}0.784 & \cellcolor[HTML]{FFCE93}0.239 &  & \cellcolor[HTML]{FFFC9E}11.87  & \cellcolor[HTML]{FFFC9E}0.462 & \cellcolor[HTML]{FFFC9E}0.834 &  & -                             & -                             & -                             &  & -                             & -                             & -                             &  & \cellcolor[HTML]{FFCCC9}25.93 & \cellcolor[HTML]{FFCCC9}0.867 & \cellcolor[HTML]{FFCCC9}0.177 \\ \hline
garden                   & \cellcolor[HTML]{FFFC9E}24.85 & \cellcolor[HTML]{FFFC9E}0.729 & \cellcolor[HTML]{FFFC9E}0.126 &  & 18.13                          & 0.281                         & 0.485                         &  & \cellcolor[HTML]{FFCE93}25.47 & \cellcolor[HTML]{FFCCC9}0.839 & \cellcolor[HTML]{FFCCC9}0.107 &  & -                             & -                             & -                             &  & \cellcolor[HTML]{FFCCC9}26.44 & \cellcolor[HTML]{FFCE93}0.82  & \cellcolor[HTML]{FFCE93}0.131 \\
counter                  & \cellcolor[HTML]{FFCE93}27.57 & \cellcolor[HTML]{FFFC9E}0.862 & \cellcolor[HTML]{FFFC9E}0.209 &  & 15.02                          & 0.537                         & 0.632                         &  & \cellcolor[HTML]{FFFC9E}26.87 & \cellcolor[HTML]{FFCE93}0.873 & \cellcolor[HTML]{FFCCC9}0.124 &  & -                             & -                             & -                             &  & \cellcolor[HTML]{FFCCC9}28.80 & \cellcolor[HTML]{FFCCC9}0.897 & \cellcolor[HTML]{FFCE93}0.157 \\
bicycle                  & \cellcolor[HTML]{FFFC9E}17.52 & \cellcolor[HTML]{FFFC9E}0.303 & \cellcolor[HTML]{FFFC9E}0.567 &  & 16.09                          & 0.256                         & 0.634                         &  & \cellcolor[HTML]{FFCE93}23.10 & \cellcolor[HTML]{FFCE93}0.707 & \cellcolor[HTML]{FFCCC9}0.201 &  & -                             & -                             & -                             &  & \cellcolor[HTML]{FFCCC9}24.89 & \cellcolor[HTML]{FFCCC9}0.727 & \cellcolor[HTML]{FFCE93}0.252 \\
room                     & \cellcolor[HTML]{FFCE93}30.66 & \cellcolor[HTML]{FFCE93}0.899 & \cellcolor[HTML]{FFCE93}0.204 &  & \cellcolor[HTML]{FFFC9E}14.06  & \cellcolor[HTML]{FFFC9E}0.563 & \cellcolor[HTML]{FFFC9E}0.709 &  & -                             & -                             & -                             &  & -                             & -                             & -                             &  & \cellcolor[HTML]{FFCCC9}31.82 & \cellcolor[HTML]{FFCCC9}0.924 & \cellcolor[HTML]{FFCCC9}0.154 \\ \hline
scan69                   & \cellcolor[HTML]{FFCE93}26.37 & \cellcolor[HTML]{FFCE93}0.865 & \cellcolor[HTML]{FFCE93}0.134 &  & 15.76                          & 0.447                         & 0.565                         &  & -                             & -                             & -                             &  & \cellcolor[HTML]{FFFC9E}18.09 & \cellcolor[HTML]{FFFC9E}0.554 & \cellcolor[HTML]{FFFC9E}0.521 &  & \cellcolor[HTML]{FFCCC9}26.62 & \cellcolor[HTML]{FFCCC9}0.868 & \cellcolor[HTML]{FFCCC9}0.112 \\
scan83                   & \cellcolor[HTML]{FFCE93}28.36 & \cellcolor[HTML]{FFCE93}0.882 & \cellcolor[HTML]{FFCE93}0.172 &  & \cellcolor[HTML]{FFFC9E}20.45 & \cellcolor[HTML]{FFFC9E}0.759 & \cellcolor[HTML]{FFFC9E}0.321 &  & -                             & -                             & -                             &  & 12.81                        & 0.572                         & 0.546                         &  & \cellcolor[HTML]{FFCCC9}28.44 & \cellcolor[HTML]{FFCCC9}0.881 & \cellcolor[HTML]{FFCCC9}0.117 \\ 
scan106                  & \cellcolor[HTML]{FFCE93}32.74 & \cellcolor[HTML]{FFCE93}0.923 & \cellcolor[HTML]{FFCE93}0.109 &  & \cellcolor[HTML]{FFFC9E}20.30  & \cellcolor[HTML]{FFFC9E}0.664 & \cellcolor[HTML]{FFFC9E}0.379 &  & -                             & -                             & -                             &  & 18.00                         & 0.550                         & 0.530                         &  & \cellcolor[HTML]{FFCCC9}34.35 & \cellcolor[HTML]{FFCCC9}0.936 & \cellcolor[HTML]{FFCCC9}0.066 \\
scan110                  & \cellcolor[HTML]{FFCE93}31.46 & \cellcolor[HTML]{FFCE93}0.905 & \cellcolor[HTML]{FFCE93}0.142 &  & \cellcolor[HTML]{FFFC9E}21.58  & \cellcolor[HTML]{FFFC9E}0.752 & \cellcolor[HTML]{FFFC9E}0.323 &  & -                             & -                             & -                             &  & 18.87                         & 0.644                         & 0.482                         &  & \cellcolor[HTML]{FFCCC9}32.63 & \cellcolor[HTML]{FFCCC9}0.931 & \cellcolor[HTML]{FFCCC9}0.074 \\
\hline\end{tabular}
    }
    \vspace{-0.3cm}
    \caption{Quantitative comparison of novel view synthesis. (-) denotes unreported results for ZeroGS and failed scenes for CF-3DGS.}    
    \label{tab: nvs}
\end{table*}
\begin{table*}
    \centering
    \vspace{-0.2cm}
    \resizebox{\textwidth}{!}{ 
\begin{tabular}{ccccccccccccccc}
\hline
& \multicolumn{2}{c}{3DGS} &  & \multicolumn{2}{c}{Spann3R}   &  & \multicolumn{2}{c}{ZeroGS}  &  & \multicolumn{2}{c}{CF-3DGS} &  & \multicolumn{2}{c}{Ours} \\ \cline{2-3} \cline{5-6} \cline{8-9} \cline{11-12} \cline{14-15} 
\multirow{-2}{*}{Scenes} & Rotation(°)$\downarrow$                  & ATE(m)$\downarrow$                         &  & Rotation(°)$\downarrow$                   & ATE(m)$\downarrow$                        &  & Rotation(°)$\downarrow$                  & ATE(m)$\downarrow$                       &  & Rotation(°)$\downarrow$    & ATE(m)$\downarrow$    &  & Rotation(°)$\downarrow$                  & ATE(m)$\downarrow$                        \\ \hline
Truck    & \cellcolor[HTML]{FFCE93}0.83 & \cellcolor[HTML]{FFCE93}0.027  &  & \cellcolor[HTML]{FFFC9E}51.69 & \cellcolor[HTML]{FFFC9E}3.156  &  & -   & -                             &  & -              & -          &  & \cellcolor[HTML]{FFCCC9}0.16 & \cellcolor[HTML]{FFCCC9}0.011 \\
Ignatius                 & \cellcolor[HTML]{FFFC9E}0.23 & \cellcolor[HTML]{FFFC9E}0.016  &  & 5.87                          & 0.391                          &  & \cellcolor[HTML]{FFCCC9}0.03 & \cellcolor[HTML]{FFCCC9}0.002 &  & -              & -          &  & \cellcolor[HTML]{FFCE93}0.06 & \cellcolor[HTML]{FFCE93}0.005 \\
Caterpillar               & \cellcolor[HTML]{FFCE93}1.41 & \cellcolor[HTML]{FFCE93}0.402   &  & \cellcolor[HTML]{FFFC9E}10.95  & \cellcolor[HTML]{FFFC9E}0.695 &  & \cellcolor[HTML]{FFFFFF}-    & -                             &  & 82.50           & 3.743      &  & \cellcolor[HTML]{FFCCC9}0.32 & \cellcolor[HTML]{FFCCC9}0.020   \\
Meetingroom              & \cellcolor[HTML]{FFCE93}0.75 & \cellcolor[HTML]{FFCE93}0.052  &  & \cellcolor[HTML]{FFFC9E}26.14 & \cellcolor[HTML]{FFFC9E}2.021  &  & \cellcolor[HTML]{FFFFFF}-    & -                             &  & -              & -          &  & \cellcolor[HTML]{FFCCC9}0.24 & \cellcolor[HTML]{FFCCC9}0.023  \\ \hline
garden                   & \cellcolor[HTML]{FFFC9E}0.19 & \cellcolor[HTML]{FFFC9E}0.003  &  & 2.08                          & 0.147                          &  & \cellcolor[HTML]{FFCCC9}0.03 & \cellcolor[HTML]{FFCCC9}0.002 &  & -              & -          &  & \cellcolor[HTML]{FFCCC9}0.03 & \cellcolor[HTML]{FFCCC9}0.002  \\
counter                  & \cellcolor[HTML]{FFFC9E}0.25 & \cellcolor[HTML]{FFFC9E}0.011  &  & 4.08                          & 0.332                          &  & \cellcolor[HTML]{FFCCC9}0.03 & \cellcolor[HTML]{FFCCC9}0.002 &  & -              & -          &  & \cellcolor[HTML]{FFCE93}0.05 & \cellcolor[HTML]{FFCE93}0.003 \\
bicycle                  & \cellcolor[HTML]{FFFC9E}1.07 & \cellcolor[HTML]{FFFC9E}0.034  &  & 11.11                         & 1.516                           &  & \cellcolor[HTML]{FFCCC9}0.04 & \cellcolor[HTML]{FFCCC9}0.005 &  & -              & -          &  & \cellcolor[HTML]{FFCE93}0.09 & \cellcolor[HTML]{FFCE93}0.013 \\
room                     & \cellcolor[HTML]{FFCE93}0.27 & \cellcolor[HTML]{FFCE93}0.016   &  & \cellcolor[HTML]{FFFC9E}8.46  & \cellcolor[HTML]{FFFC9E}0.908  &  & -                            & -                             &  & -              & -          &  & \cellcolor[HTML]{FFCCC9}0.13 & \cellcolor[HTML]{FFCCC9}0.012 \\ \hline
scan69                   & \cellcolor[HTML]{FFCE93}0.23 & \cellcolor[HTML]{FFCE93}0.006  &  & \cellcolor[HTML]{FFFC9E}5.10   & \cellcolor[HTML]{FFFC9E}0.158  &  & -                            & -                             &  & 47.95          & 0.955      &  & \cellcolor[HTML]{FFCCC9}0.1  & \cellcolor[HTML]{FFCCC9}0.003 \\
scan83                   & \cellcolor[HTML]{FFCE93}0.26 & \cellcolor[HTML]{FFCE93}0.007 &  & \cellcolor[HTML]{FFFC9E}3.04  & \cellcolor[HTML]{FFFC9E}0.184  &  & -                            & -                             &  & 155.34         & 1.286      &  & \cellcolor[HTML]{FFCCC9}0.19 & \cellcolor[HTML]{FFCCC9}0.005 \\
scan106                  & \cellcolor[HTML]{FFCE93}0.13 & \cellcolor[HTML]{FFCE93}0.004  &  & \cellcolor[HTML]{FFFC9E}4.09  & \cellcolor[HTML]{FFFC9E}0.121  &  & -                            & -                             &  & 46.40           & 0.902      &  & \cellcolor[HTML]{FFCCC9}0.11 & \cellcolor[HTML]{FFCCC9}0.003  \\
scan110                  & \cellcolor[HTML]{FFCE93}0.48 & \cellcolor[HTML]{FFCE93}0.007 &  & \cellcolor[HTML]{FFFC9E}3.00     & \cellcolor[HTML]{FFFC9E}0.129  &  & -                            & -                             &  & 66.78          & 0.983      &  & \cellcolor[HTML]{FFCCC9}0.13 & \cellcolor[HTML]{FFCCC9}0.004 \\
\hline
\end{tabular}
    }
    \vspace{-0.3cm}
    \caption{Quantitative comparison of camera pose registration. (-) indicates unreported results for ZeroGS and failed scenes for CF-3DGS.}    \label{tab: pose}
    \vspace{-0.5cm}
\end{table*}
\begin{figure*}[t]
    \centering
    \vspace{ -0.6cm}
    \includegraphics[width=\textwidth]{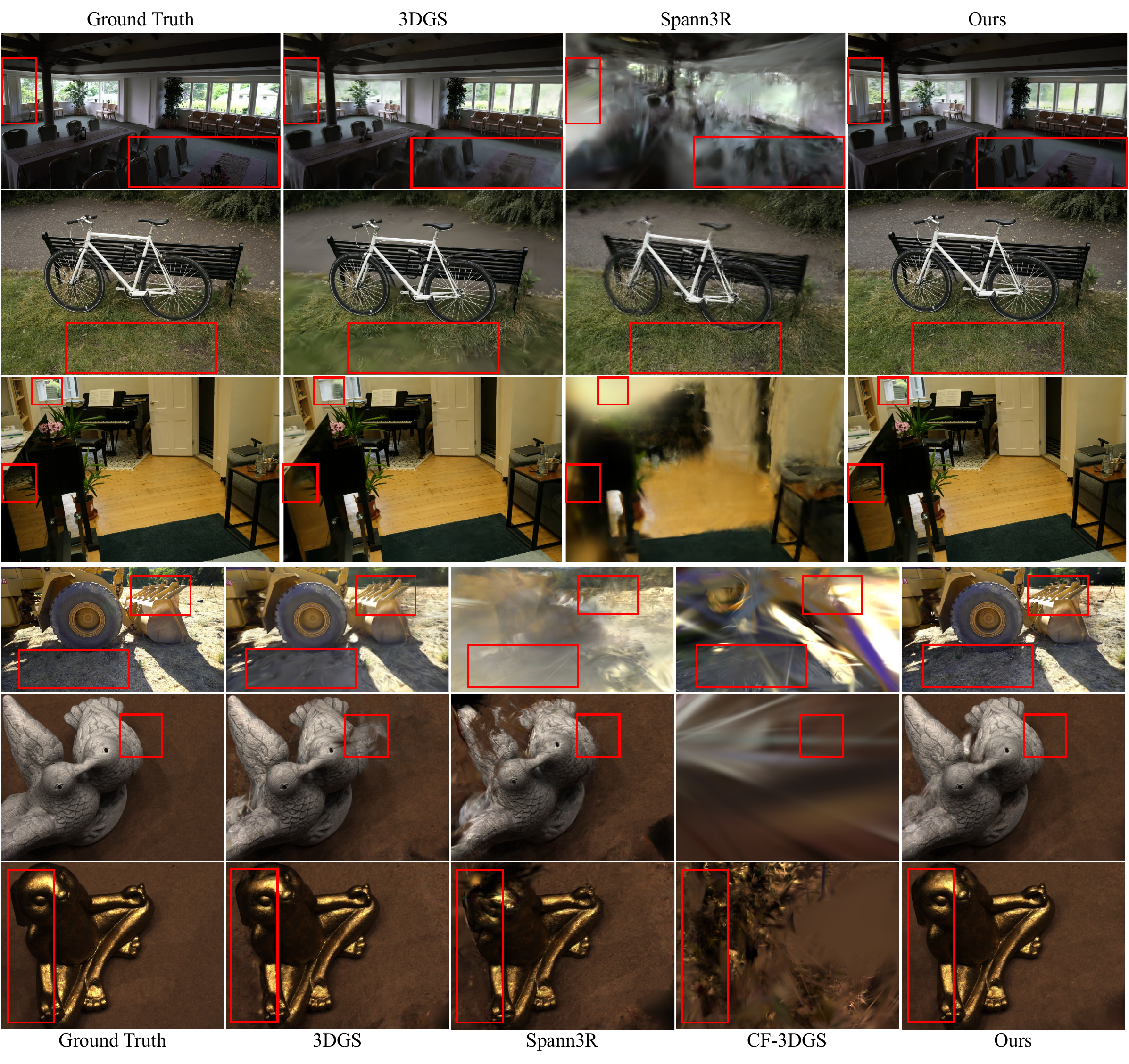}
    \vspace{-0.9cm}
    \caption{Results for novel view synthesis. We omit the failure scenes for CF-3DGS and unreported results for ZeroGS.}
    \label{fig:visual}
\end{figure*}

\begin{figure*}[h]
    \centering
    \vspace{ -0.6cm}
    \includegraphics[width=\textwidth]{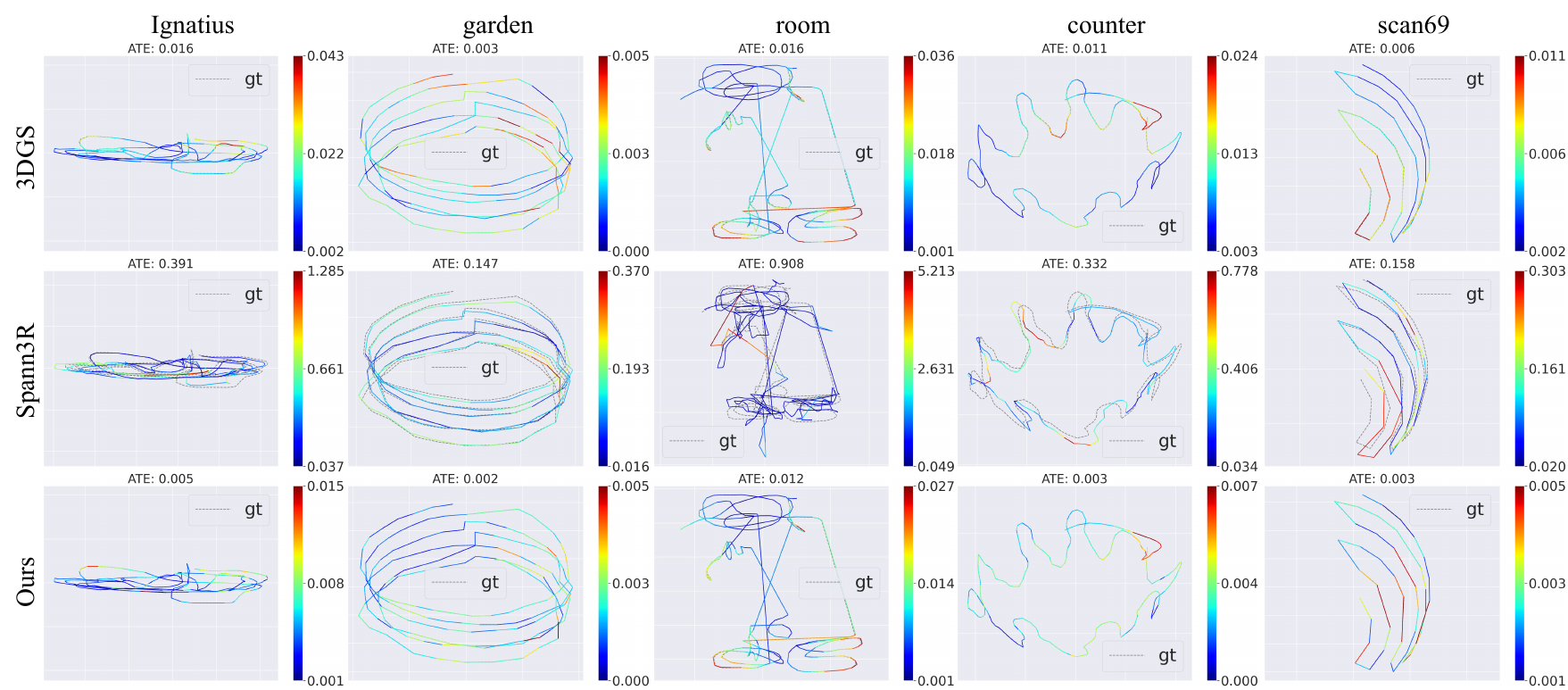}
    \vspace{ -0.7cm}
    \caption{Visualization of camera pose registration for the three best-performing methods. ZeroGS results are unavailable.}
    \label{fig:visual-pose}
\end{figure*}

\begin{table*}[h]
    \centering
    \vspace{ -0.2cm}
    \resizebox{\textwidth}{!}{ 
\begin{tabular}{cccclccclccclccclccc}
\hline
                         & \multicolumn{3}{c}{BARF}                                                                    &                          & \multicolumn{3}{c}{SC-NeRF}                                                                 &  & \multicolumn{3}{c}{Nope-NeRF}                                                               &  & \multicolumn{3}{c}{CF-3DGS}                                                                 &  & \multicolumn{3}{c}{Ours}                                                                    \\ \cline{2-4} \cline{6-8} \cline{10-12} \cline{14-16} \cline{18-20} 
\multirow{-2}{*}{Scenes} & PSNR$\uparrow$                          & SSIM$\uparrow$                         & LPIPS$\downarrow$                        &                          & PSNR$\uparrow$                          & SSIM$\uparrow$                         & LPIPS$\downarrow$                        &  & PSNR$\uparrow$                          & SSIM$\uparrow$                         & LPIPS$\downarrow$                        &  & PSNR$\uparrow$                          & SSIM$\uparrow$                         & LPIPS$\downarrow$                        &  & PSNR$\uparrow$                          & SSIM$\uparrow$                         & LPIPS$\downarrow$                        \\ \hline
Barn                     & \cellcolor[HTML]{FFFFFF}25.28 & \cellcolor[HTML]{FFFFFF}0.64 & \cellcolor[HTML]{FFFFFF}0.48 & \cellcolor[HTML]{FFFFFF} & \cellcolor[HTML]{FFFFFF}23.26 & \cellcolor[HTML]{FFFFFF}0.62 & \cellcolor[HTML]{FFFFFF}0.51 &  & \cellcolor[HTML]{FFFC9E}26.35 & \cellcolor[HTML]{FFFC9E}0.69 & \cellcolor[HTML]{FFFC9E}0.44 &  & \cellcolor[HTML]{FFCE93}31.23 & \cellcolor[HTML]{FFCE93}0.93 & \cellcolor[HTML]{FFCE93}0.11 &  & \cellcolor[HTML]{FFCCC9}36.95 & \cellcolor[HTML]{FFCCC9}0.97 & \cellcolor[HTML]{FFCCC9}0.01 \\
Museum                   & \cellcolor[HTML]{FFFFFF}23.58 & \cellcolor[HTML]{FFFFFF}0.61 & \cellcolor[HTML]{FFFFFF}0.55 & \cellcolor[HTML]{FFFFFF} & \cellcolor[HTML]{FFFFFF}24.94 & \cellcolor[HTML]{FFFFFF}0.69 & \cellcolor[HTML]{FFFFFF}0.45 &  & \cellcolor[HTML]{FFFC9E}26.77 & \cellcolor[HTML]{FFFC9E}0.76 & \cellcolor[HTML]{FFFC9E}0.35 &  & \cellcolor[HTML]{FFCE93}29.91 & \cellcolor[HTML]{FFCE93}0.91 & \cellcolor[HTML]{FFCE93}0.11 &  & \cellcolor[HTML]{FFCCC9}35.78 & \cellcolor[HTML]{FFCCC9}0.97 & \cellcolor[HTML]{FFCCC9}0.01 \\
Ballroom                 & \cellcolor[HTML]{FFFFFF}20.66 & \cellcolor[HTML]{FFFFFF}0.50  & \cellcolor[HTML]{FFFFFF}0.60  & \cellcolor[HTML]{FFFFFF} & \cellcolor[HTML]{FFFFFF}22.64 & \cellcolor[HTML]{FFFFFF}0.61 & \cellcolor[HTML]{FFFFFF}0.48 &  & \cellcolor[HTML]{FFFC9E}25.33 & \cellcolor[HTML]{FFFC9E}0.72 & \cellcolor[HTML]{FFFC9E}0.38 &  & \cellcolor[HTML]{FFCE93}32.47 & \cellcolor[HTML]{FFCE93}0.96 & \cellcolor[HTML]{FFCE93}0.07 &  & \cellcolor[HTML]{FFCCC9}34.56 & \cellcolor[HTML]{FFCCC9}0.97 & \cellcolor[HTML]{FFCCC9}0.01 \\
Ignatius                 & \cellcolor[HTML]{FFFFFF}21.78 & \cellcolor[HTML]{FFFFFF}0.47 & \cellcolor[HTML]{FFFFFF}0.60  & \cellcolor[HTML]{FFFFFF} & \cellcolor[HTML]{FFFFFF}23.00 & \cellcolor[HTML]{FFFFFF}0.55 & \cellcolor[HTML]{FFFFFF}0.53 &  & \cellcolor[HTML]{FFFC9E}23.96 & \cellcolor[HTML]{FFFC9E}0.61 & \cellcolor[HTML]{FFFC9E}0.47 &  & \cellcolor[HTML]{FFCE93}28.43 & \cellcolor[HTML]{FFCE93}0.90  & \cellcolor[HTML]{FFCE93}0.09 &  & \cellcolor[HTML]{FFCCC9}31.16 & \cellcolor[HTML]{FFCCC9}0.93 & \cellcolor[HTML]{FFCCC9}0.03 \\ \hline
\end{tabular}
    }
    \vspace{-0.3cm}
    \caption{Quantitative comparison of novel view synthesis on short video clips.}
    \vspace{ -0.7cm}
    \label{tab: clips}
\end{table*}

\begin{table*}[h]
    \centering
    \vspace{ -0.2cm}
    \resizebox{\textwidth}{!}{ 

\begin{tabular}{cccclccclccclccclccc}
\hline
                         & \multicolumn{3}{c}{BARF}    &  & \multicolumn{3}{c}{SC-NeRF}                                                                   &  & \multicolumn{3}{c}{Nope-NeRF}                                                                 &  & \multicolumn{3}{c}{CF-3DGS}                                                                   &  & \multicolumn{3}{c}{Ours}                                                                      \\ \cline{2-4} \cline{6-8} \cline{10-12} \cline{14-16} \cline{18-20} 
\multirow{-2}{*}{scenes} & \( \mathrm{RPE}_t \downarrow \) & \( \mathrm{RPE}_r \downarrow \) & ATE $\downarrow$   &  & \( \mathrm{RPE}_t \downarrow \)                      & \( \mathrm{RPE}_r \downarrow \)                      & ATE $\downarrow$                           &  & \( \mathrm{RPE}_t \downarrow \)                      & \( \mathrm{RPE}_r \downarrow \)                      & ATE $\downarrow$                           &  & \( \mathrm{RPE}_t \downarrow \)                  & \( \mathrm{RPE}_r \downarrow \)                  & ATE $\downarrow$                           &  & \( \mathrm{RPE}_t \downarrow \)                      & \( \mathrm{RPE}_r \downarrow \)                      & ATE $\downarrow$                           \\ \hline
Barn                     & 0.314    & 0.265    & 0.050  &  & \cellcolor[HTML]{FFFFFF}1.317 & \cellcolor[HTML]{FFFFFF}0.429 & \cellcolor[HTML]{FFFFFF}0.157 &  & \cellcolor[HTML]{FFFC9E}0.046 & \cellcolor[HTML]{FFCE93}0.032 & \cellcolor[HTML]{FFFC9E}0.004 &  & \cellcolor[HTML]{FFCE93}0.034 & \cellcolor[HTML]{FFFC9E}0.034 & \cellcolor[HTML]{FFCE93}0.003 &  & \cellcolor[HTML]{FFCCC9}0.009 & \cellcolor[HTML]{FFCCC9}0.020  & \cellcolor[HTML]{FFCCC9}0.000     \\
Museum                   & 3.442    & 1.128    & 0.263 &  & \cellcolor[HTML]{FFFFFF}8.339 & \cellcolor[HTML]{FFFFFF}1.491 & \cellcolor[HTML]{FFFFFF}0.316 &  & \cellcolor[HTML]{FFFC9E}0.207 & \cellcolor[HTML]{FFCE93}0.202 & \cellcolor[HTML]{FFFC9E}0.020  &  & \cellcolor[HTML]{FFCE93}0.052 & \cellcolor[HTML]{FFFC9E}0.215 & \cellcolor[HTML]{FFCE93}0.005 &  & \cellcolor[HTML]{FFCCC9}0.018 & \cellcolor[HTML]{FFCCC9}0.020  & \cellcolor[HTML]{FFCCC9}0.000     \\
Ballroom                 & 0.531    & 0.228    & 0.018 &  & \cellcolor[HTML]{FFFFFF}0.328 & \cellcolor[HTML]{FFFFFF}0.146 & \cellcolor[HTML]{FFFFFF}0.012 &  & \cellcolor[HTML]{FFFC9E}0.041 & \cellcolor[HTML]{FFCE93}0.018 & \cellcolor[HTML]{FFCE93}0.002 &  & \cellcolor[HTML]{FFCE93}0.037 & \cellcolor[HTML]{FFFC9E}0.024 & \cellcolor[HTML]{FFFC9E}0.003 &  & \cellcolor[HTML]{FFCCC9}0.016 & \cellcolor[HTML]{FFCCC9}0.013 & \cellcolor[HTML]{FFCCC9}0.000     \\
Ignatius                 & 0.736    & 0.324    & 0.029 &  & \cellcolor[HTML]{FFFFFF}0.533 & \cellcolor[HTML]{FFFFFF}0.240  & \cellcolor[HTML]{FFFFFF}0.085 &  & \cellcolor[HTML]{FFCE93}0.026 & \cellcolor[HTML]{FFCE93}0.005 & \cellcolor[HTML]{FFCE93}0.002 &  & \cellcolor[HTML]{FFFC9E}0.033 & \cellcolor[HTML]{FFFC9E}0.032 & \cellcolor[HTML]{FFFC9E}0.005 &  & \cellcolor[HTML]{FFCCC9}0.010  & \cellcolor[HTML]{FFCCC9}0.012 & \cellcolor[HTML]{FFCCC9}0.001 \\ \hline
\end{tabular}
    }
    \vspace{ -0.3cm}
    \caption{Quantitative comparison of camera pose registration on short video clips.}
    \vspace{-0.5cm}
    \label{tab: cfgs-pose}
\end{table*}
\begin{table}[h]
    \centering
    \resizebox{\columnwidth}{!}{ 

\begin{tabular}{cccclccc}
\hline
                         & \multicolumn{3}{c}{COLMAP}                                 &  & \multicolumn{3}{c}{Ours}                                                                       \\ \cline{2-4} \cline{6-8} 
                         & \multicolumn{2}{c}{Pose}               &                        &  & \multicolumn{2}{c}{Pose}                                      &                                \\ \cline{2-3} \cline{6-7}
\multirow{-3}{*}{Scenes} & R(°)$\downarrow$   & ATE(m)$\downarrow$                        & \multirow{-2}{*}{PSNR$\uparrow$} &  & R(°)$\downarrow$                         & ATE(m)$\downarrow$                       & \multirow{-2}{*}{PSNR$\uparrow$}         \\ \hline
0079\_00                 & 3.55   & \cellcolor[HTML]{FFCCC9}0.014 & 30.78                 &  & \cellcolor[HTML]{FFCCC9}2.45  & \cellcolor[HTML]{FFCCC9}0.014 & \cellcolor[HTML]{FFCCC9}32.58  \\
0301\_00                 & 133.83 & 0.169                         & 23.63                 &  & \cellcolor[HTML]{FFCCC9}9.30   & \cellcolor[HTML]{FFCCC9}0.009 & \cellcolor[HTML]{FFCCC9}30.11  \\
0418\_00                 & 5.03   & \cellcolor[HTML]{FFCCC9}0.012 & 29.03                 &  & \cellcolor[HTML]{FFCCC9}4.34  & \cellcolor[HTML]{FFCCC9}0.012 & \cellcolor[HTML]{FFCCC9}31.62 \\
\hline
\end{tabular}
    }
    \vspace{ -0.3cm}
    \caption{Comparison with COLMAP on ScanNet.}
    \label{tab: colmap}
\end{table}
\begin{table}
    \centering
    \vspace{-0.3cm}
    \resizebox{\columnwidth}{!}{ 

\begin{tabular}{cccclcclccc}
\hline
\multicolumn{4}{c}{configs}                                                                                   &  & \multicolumn{2}{c}{Pose}                                      &  & \multicolumn{3}{c}{NVS}                                                                        \\ \cline{1-4} \cline{6-7} \cline{9-11} 
3DGS                      & MCMC                      & $\mathcal{R}_{\text{MLP}}$              & $\mathcal{L}_{\text {geo }}$                   &  & R(°)$\downarrow$                          & ATE(m)$\downarrow$                        &  & PSNR$\uparrow$                           & SSIM$\uparrow$                          & LPIPS$\downarrow$                         \\ \hline
\checkmark &                           &                           &                           &  & 0.805                         & 0.124                         &  & 20.49                         & 0.678                         & 0.255                         \\
\checkmark & \checkmark &                           &                           &  & \cellcolor[HTML]{FFFC9E}0.783 & \cellcolor[HTML]{FFFC9E}0.032 &  & \cellcolor[HTML]{FFFC9E}21.94 & \cellcolor[HTML]{FFFC9E}0.724 & \cellcolor[HTML]{FFFC9E}0.222 \\
\checkmark & \checkmark & \checkmark &                           &  & \cellcolor[HTML]{FFCE93}0.353 & \cellcolor[HTML]{FFCE93}0.025 &  & \cellcolor[HTML]{FFCE93}23.61 & \cellcolor[HTML]{FFCE93}0.803 & \cellcolor[HTML]{FFCE93}0.193 \\
\checkmark & \checkmark & \checkmark & \checkmark &  & \cellcolor[HTML]{FFCCC9}0.195 & \cellcolor[HTML]{FFCCC9}0.015 &  & \cellcolor[HTML]{FFCCC9}24.02 & \cellcolor[HTML]{FFCCC9}0.820 & \cellcolor[HTML]{FFCCC9}0.183 \\ \hline

\multicolumn{4}{c}{Geometric Constraints Only }                        &  & 0.430                          & 0.038                         &  & 21.94                         & 0.731                         & 0.239                       \\ 
\multicolumn{4}{c}{Local Geometric Constraints}                       &  & 0.458                       & 0.034                      &  & 21.97                         & 0.736                         & 0.228                         \\ \hline
\end{tabular}
    }
    \vspace{ -0.3cm}
    \caption{Ablation Study on Tanks and Templates.}
    \label{tab:ablation}
    \vspace{-0.6cm}
\end{table}

\subsection{Experimental Setup}
\label{sec:setup}
\noindent \textbf{Datasets.} We evaluate our method on three widely used real-world datasets: Tanks and Temples \cite{knapitsch2017tanks}, Mip-NeRF360 \cite{barron2022mip}, and DTU \cite{jensen2014large}, selecting four representative scenes from each.
The Mip-NeRF360 dataset contains indoor and outdoor scenes captured with cameras distributed evenly along 360-degree trajectories, with each scene comprising approximately 100–300 images. Tanks and Temples follows a similar setup in terms of camera poses and scene scale but exhibits greater variations in illumination and appearance. In contrast, DTU focuses on object-level indoor scenes captured under controlled lighting, with each sequence containing 49 or 64 images and precise ground truth poses.
These datasets are widely used in the 3D Gaussian Splatting literature \cite{kerbl3Dgaussians, huang20242d}. Following prior work, we adopt the same evaluation protocol and training view resolution for each dataset.

\noindent \textbf{Metrics.} 
Following BARF \cite{lin2021barf} and CF-3DGS \cite{fu2024colmap}, we evaluate both Novel View Synthesis (NVS) and camera pose registration. For camera pose evaluation, we report the average rotation error and the Root Mean Square Error (RMSE) of the Absolute Trajectory Error (ATE) \cite{matsuki2024gaussian} (in meters) on the training views. To account for similarity transformations, we align the optimized training poses with the ground truth using Procrustes analysis on camera locations, following prior work \cite{lin2021barf}.
For NVS, we report PSNR, SSIM \cite{wang2004image}, and LPIPS \cite{zhang2018unreasonable}. Since NVS requires test view poses, we perform test-time rendering optimization to obtain optimal test poses, consistent with previous approaches \cite{lin2021barf, fu2024colmap, bian2023nope}.

\noindent \textbf{Implementation details.} 
Our method is implemented in PyTorch, building upon the 3D Gaussian Splatting framework gsplat \cite{ye2024gsplatopensourcelibrarygaussian}.
For all experiments, we employ consistent weighting factors:
$\lambda_{\text{D-SSIM}}=0.2$, $\lambda_{o}=0.01$, $\lambda_{\Sigma}=0.01$, and $\lambda_{\text{geo}}=2$.
We find that the epipolar geometric constraint $\mathcal{L}_{\text {geo }}$ plays a crucial role in the early training phase of 3DGS, helping to establish correct geometry, its influence becomes less critical in later stages. Thus we decay $\lambda_{\text{geo}}$ to $0$ after 3,000 iterations. This strategy is consistently applied across all our experiments.
Our training procedure follows the standard 3DGS pipeline \cite{kerbl3Dgaussians,huang20242d}, where each iteration samples and renders a single training view.
The key distinction is that we additionally compute $\mathcal{L}_{\text {geo }}$ at each step and propagate camera pose gradients through our pose refiner and associated camera latent codes.
All experiments are conducted on a single NVIDIA RTX 4090 GPU with 24GB memory.

\subsection{Comparison with Previous Methods}
\noindent \textbf{Results on full video sequence.}
We evaluate our method against four state-of-the-art baselines: 3DGS \cite{kerbl3Dgaussians}, Spann3R \cite{wang20243d}, ZeroGS \cite{chen2024zerogs}, and CF-3DGS \cite{fu2024colmap}. The evaluation is conducted on three standard datasets: Tanks and Temples \cite{knapitsch2017tanks}, Mip-NeRF360 \cite{barron2022mip}, and DTU \cite{jensen2014large}.
For 3DGS, we utilize the gsplat implementation \cite{ye2024gsplatopensourcelibrarygaussian}. To ensure fair comparison, we configure 3DGS to use the same camera poses obtained from MASt3R-SfM and enable camera pose optimization during training, as supported by gsplat \cite{ye2024gsplatopensourcelibrarygaussian}.
Since Spann3R employs a different scene representation, we use its estimated camera poses for 3DGS training with camera optimization enabled, maintaining consistency with our experimental setup. For ZeroGS, which lacks publicly available code at the time of writing, we report results directly from their paper. CF-3DGS experiments are conducted using their official implementation across all three datasets.

Table~\ref{tab: nvs} and Fig.~\ref{fig:visual} present the novel view synthesis results. 
Our approach demonstrates superior performance across all three datasets, significantly outperforming 3DGS, which uses identical MASt3R-SfM camera poses. 
While ZeroGS shows slightly lower performance compared to our method, CF-3DGS consistently fails on Mip-NeRF360 and Tanks and Temples datasets. 
This failure can be attributed to early camera tracking loss during their progressive training pipeline, particularly evident in scenes with large camera motion.

For camera registration (Table~\ref{tab: pose} and Fig.~\ref{fig:visual-pose}), our method significantly outperforms 3DGS with camera pose optimization, while showing comparable results to ZeroGS. 
The marginal difference (0.02° in rotation error and 0.003 m in ATE on average) is negligible, especially considering our superior novel view synthesis results. 
Moreover, while ZeroGS employs a complex two-stage training strategy with progressive image registration similar to classical incremental SfM, our approach achieves competitive results through a simpler process that enhances standard 3DGS training using SfM outputs, introducing minimal computational overhead.

\noindent \textbf{Results on short video clips.}
We also evaluate our method on short video clips from the Tanks and Temples dataset, following the experimental protocol of CF-3DGS\cite{fu2024colmap} and using their preprocessed data.
For comparison, we select state-of-the-art baselines including BARF \cite{lin2021barf}, SC-NeRF \cite{jeong2021self}, and Nope-NeRF \cite{bian2023nope}.

As shown in Table~\ref{tab: clips} and Table~\ref{tab: cfgs-pose}, our approach demonstrates substantial improvements over existing methods in both novel view synthesis and camera pose estimation. 
For fair comparison, we follow the evaluation protocol from prior work, where the Absolute Trajectory Error (ATE) is scaled by a factor of 100, and camera pose alignment with ground truth is performed using both translation and rotation. 
All quantitative results are obtained using the official evaluation code provided by \cite{fu2024colmap}.

\noindent \textbf{Comparison with COLMAP.}
While both the Tanks and Temples dataset and MipNeRF360 use COLMAP-generated camera poses as ground truth due to its generally high accuracy, we identify specific scenarios where COLMAP faces challenges. 
To demonstrate this, we evaluate several challenging scenes from the ScanNet \cite{dai2017scannet} dataset, with results presented in Table~\ref{tab: colmap}. Our method demonstrates superior performance compared to COLMAP on these scenes, highlighting the limitations of traditional structure-from-motion approaches in challenging scenarios.

\subsection{Ablation Studies}
\noindent \textbf{Component-wise analysis.}
Our method comprises three key components: 1) 3DGS as MCMC, 2) rendering-free global geometric constraints, and 3) a correlation-modeling global camera pose refiner.
To evaluate each component's contribution, we conduct ablation studies as shown in Table~\ref{tab:ablation}.
We evaluate on scenes from the Tanks and Temples dataset (Table~\ref{tab: nvs}), reporting average metrics for both novel view synthesis and camera pose estimation.
The results demonstrate significant improvements from each component.
While the baseline 3DGS method includes camera pose optimization during training (provided by gsplat \cite{ye2024gsplatopensourcelibrarygaussian}), its effectiveness is limited without our proposed components.
Detailed per-scene ablation results for Tables~\ref{tab: nvs} and~\ref{tab: pose} are provided in the supplementary materials.

\noindent \textbf{Synergistic effects of photometric and geometric losses.}
While Table~\ref{tab:ablation} validates the effectiveness of our geometric constraints $\mathcal{L}_{\text{geo}}$, we further investigate the role of photometric rendering in camera pose registration and NVS.
We conduct an experiment by detaching camera pose parameters during gaussian splitting, eliminating gradients from the photometric rendering to camera parameters.
Additionally, we maintain $\lambda_{\text{geo}}$ constant instead of decaying it to 0 after step 3000 as described in Section~\ref{sec:setup}.
Results in the second-to-last row of Table~\ref{tab:ablation} demonstrate that the rendering loss in Eq.~\ref{eq:total} is crucial for both high-quality NVS and accurate camera pose registration.
The optimal performance is achieved through the synergistic combination of rendering loss and geometric constraints.

\noindent \textbf{Advantages of global over Local geometric constraints.}
To demonstrate the superiority of global geometric constraints over local alternatives, we compare against a variant that randomly samples each image pair for correspondence at every step, rather than utilizing all pairs ${(n, m) \in \mathcal{E}}$.
The results, shown in the last row of Table~\ref{tab:ablation}, indicate that local geometric constraints provide minimal benefit compared to our global approach, validating the effectiveness of $\mathcal{L}_{\text{geo}}$.
\section{Conclusion}
We present 3R-GS, a robust framework for optimizing camera poses and 3D Gaussian representations from imperfect MASt3R-SfM outputs.
Experimental results demonstrate our 3R-GS achieves superior performance in both novel view synthesis and camera pose registration compared to prior work.
We hope this will benefit 3R-based \cite{leroy2024grounding,wang2024dust3r} methods and their downstream application with 3DGS in community.
Possible future work could explore extending our approach to handle dynamic scenes and support real-time applications.

{
    \small
    \bibliographystyle{ieeenat_fullname}
    \bibliography{main.bib}
}

\end{document}